\documentclass[letterpaper, 10 pt, conference]{ieeeconf}  

\IEEEoverridecommandlockouts                              

\usepackage{amsmath} 
\usepackage{graphicx}
\usepackage{kotex}
\usepackage{multirow}
\usepackage{booktabs}

\usepackage{tikz}

\newcommand{\arxivcopyright}{
\begin{tikzpicture}[remember picture,overlay]
\node[anchor=south, yshift=10pt] at (current page.south) {
    \parbox{\textwidth}{
        \centering \scriptsize 
        \copyright~2026 IEEE. Personal use of this material is permitted. Permission from IEEE must be obtained for all other uses, in any current or future media, including reprinting/republishing this material for advertising or promotional purposes, creating new collective works, for resale or redistribution to servers or lists, or reuse of any copyrighted component of this work in other works.
    }
};
\end{tikzpicture}
}

\title{\LARGE \bf
TactiVerse: Generalizing Multi-Point Tactile Sensing in Soft Robotics Using Single-Point Data
}

\author{Junhui Lee$^{1,2\textsuperscript{\textdagger}}$, Hyosung Kim$^{1,2\textsuperscript{\textdagger}}$, Janghoon Choi$^1$, and Saekwang Nam$^{1*}$
\thanks{$^1$All authors are with the Graduate School of Data Science at Kyungpook National University, 41566 Daegu, Republic of Korea. 
       {\tt\small \{jk06033, hyosungk98, jhchoi09, s.nam\}@knu.ac.kr}}%
\thanks{$^2$J.L. and H.K. are enrolled in the Major in Data Science Convergence.}
\thanks{$^{*}$Corresponding author: Saekwang Nam (ORCID: 0000-0002-7713-8505)}%
\thanks{\textsuperscript{\textdagger} Denotes equal contribution.}
}

\begin{document}

\maketitle
\arxivcopyright 
\thispagestyle{empty}
\pagestyle{empty}

\begin{abstract}
Real-time prediction of deformation in highly compliant soft materials remains a significant challenge in soft robotics. While vision-based soft tactile sensors can track internal marker displacements, learning-based models for 3D contact estimation heavily depend on their training datasets, inherently limiting their ability to generalize to complex scenarios such as multi-point sensing. To address this limitation, we introduce TactiVerse, a U-Net-based framework that formulates contact geometry estimation as a spatial heatmap prediction task. Even when trained exclusively on a limited dataset of single-point indentations, our architecture achieves highly accurate single-point sensing, yielding a superior mean absolute error of 0.0589\,mm compared to the 0.0612\,mm of a conventional regression-based CNN baseline. Furthermore, we demonstrate that augmenting the training dataset with multi-point contact data substantially enhances the sensor's multi-point sensing capabilities, significantly improving the overall mean MAE for two-point discrimination from 1.214\,mm to 0.383\,mm. By successfully extrapolating complex contact geometries from fundamental interactions, this methodology unlocks advanced multi-point and large-area shape sensing. Ultimately, it significantly streamlines the development of marker-based soft sensors, offering a highly scalable solution for real-world tactile perception.

\end{abstract}

\section{INTRODUCTION}

Real-time prediction of the deformation of highly compliant soft materials upon external contact remains a significant challenge in soft robotics~\cite{Thuruthel19SciRobo_SoftRobot}. While traditional methods utilize embedded electrode array networks to infer the deformation of a soft material by reading the change of electrical signals, the inherent rigidity of this network can compromise the material's overall compliance~\cite{fan2014fractal, Mun18ToH_Wearable}. To address this, alternative strategies have emerged, such as fabricating the material to be inherently conductive~\cite{Lee23TASE_EIT, Park20ICRA_EIT}. This approach, which eliminates rigid connecting traces and relies on measuring electric potential variations between discrete points, is particularly effective for capturing deformation across large-area continuous structures~\cite{Kim24SciRobo_deformation}.


For applications requiring deformation sensing within a localized region, vision-based approaches are commonly employed to analyze the displacement of marker arrays. Prominent examples include the measurement of shear and slip at the contact interface of sensors like FingerVision or GelSight~\cite{Yamaguchi17Humanoids_Fingervision, Yuan15ICRA_GelsightMarker}. In these configurations, a camera, typically fixed behind the contact surface, tracks markers embedded within the compliant layer over time to estimate the deformation.


\begin{figure}[t!]
    \begin{center}
    \includegraphics[width = 1\columnwidth]{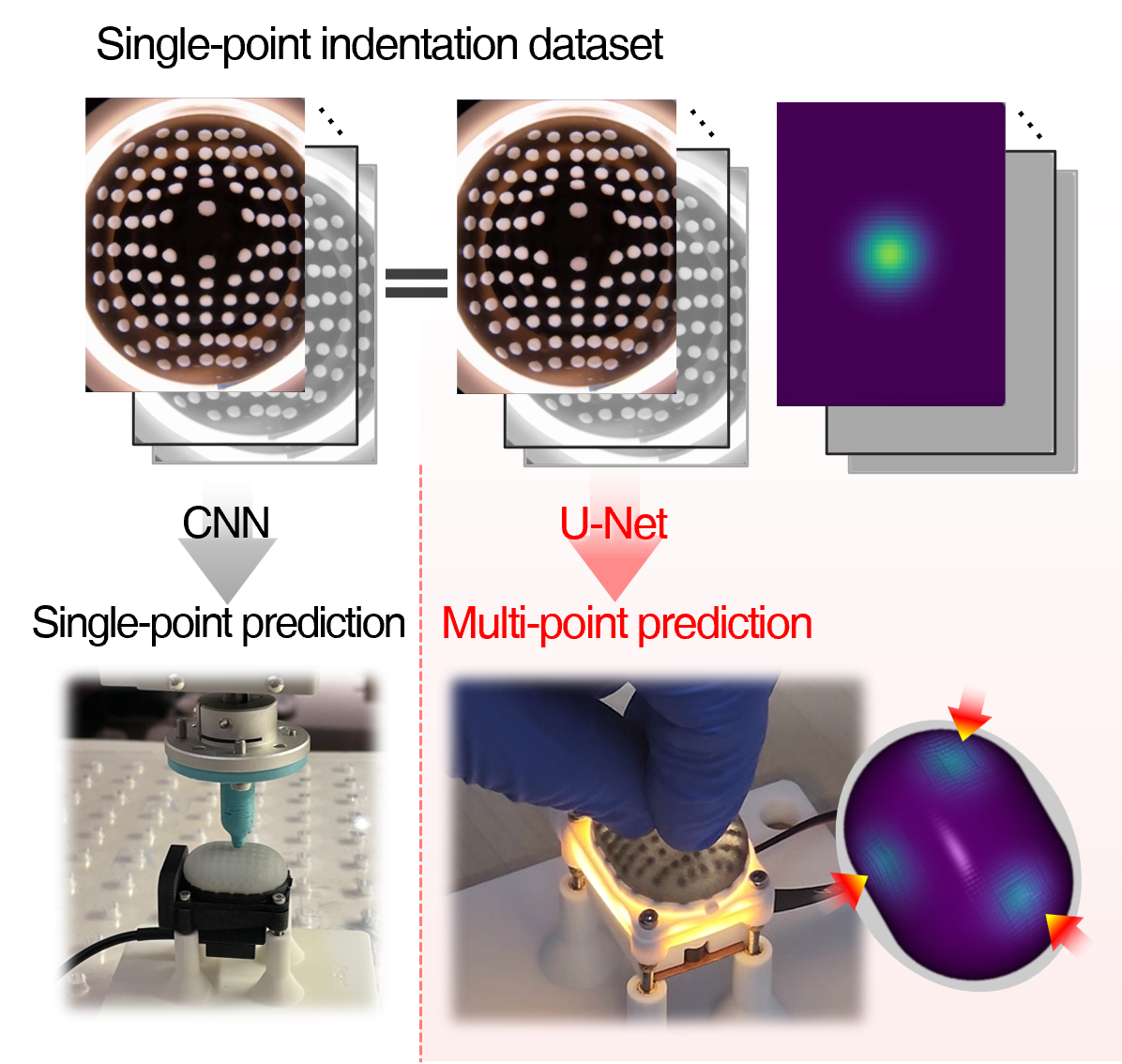}
    \caption{A comparison of a traditional convolutional neural network approach for single-point prediction and a U-Net (TactiVerse) that extends to multi-point prediction, both utilizing the same single-point indentation dataset from a TacTip soft sensor, with the U-Net employing an additional heatmap representation.}
    \label{Intro}
    \end{center}
\end{figure}

However, inferring three-dimensional (3D) deformation from 2D marker displacements is non-trivial. This challenge stems from the difficulties in achieving robust, error-free marker tracking in real-time. Furthermore, even with accurate 2D data, precise 3D reconstruction often necessitates strong assumptions, such as material incompressibility and a well-characterized mechanical model. Consequently, some sensors, like GelSight, employ alternative principles (\textit{e.g.}, photometric stereo) for 3D measurements. In contrast, other work on TacTip has utilized learning-based methods, such as applying Graph Neural Networks (GNNs) to Voronoi features derived from marker motion, to estimate 3D contact states~\cite{Fan23arXiv_VGNN}.



Furthermore, 3D deformation estimation based on supervised learning, including GNNs, is heavily dependent on the composition of the training dataset. This reliance inherently limits the ability to generalize to diverse contact geometries or complex scenarios, such as multi-point sensing. Specifically, successful inference cannot be guaranteed for contact patterns unobserved during training. While alternative studies, such as those on the BioTacTip~\cite{Li24RAL_BioTacTip}, have enabled multi-point sensing by exploiting mechanical deformation mechanisms, their practicality is often hindered by significant constraints. These, for example, include the requirement for a planar sensing surface and the inherent difficulty in precisely tuning the material properties of the contact interface.



To address these limitations, this paper introduces TactiVerse. We define TactiVerse as a U-Net framework~\cite{Ronneberger15} that effectively generates complex multi-point sensing scenarios, even when trained only on a limited dataset of single-point indentations (Fig.~\ref{Intro}). This unique capability suggests a strong potential for generalizing to complex contact patterns, including those from objects with diverse and previously unobserved geometries.


The \textbf{key contributions} of this paper are listed as follows.
\begin{itemize}
    \item We present a methodology addressing a key challenge in learning-based soft sensors: the inherent dependency on limited types of training data. Our approach enables data-driven contact geometry estimation to successfully generalize to complex contact patterns, even when trained exclusively on constrained, homogenous datasets (such as single-point indentations).
    \item Our framework expands the sensor's functional scope (\textit{e.g.}, enabling multi-point or complex contact pattern sensing) without necessitating auxiliary hardware or imposing rigid constraints near the contact interface. This approach fully preserves the intrinsic compliance and morphological advantages of the soft sensor.
\end{itemize}




\section{Materials and Methods}
This section details the methodology used in our study. The Materials subsection describes the experimental setup designed for data collection, specifically for acquiring the input-output data pairs: the contact conditions of a single-point indenter and the corresponding marker array images captured by the TacTip's internal camera. Following this, the Methods subsection elaborates on the U-Net architecture, which utilizes this single-point indentation dataset to enable the estimation of multi-point contact scenarios.


\subsection{Materials}
\begin{figure}[t!]
    \begin{center}
    \includegraphics[width = 1\columnwidth]{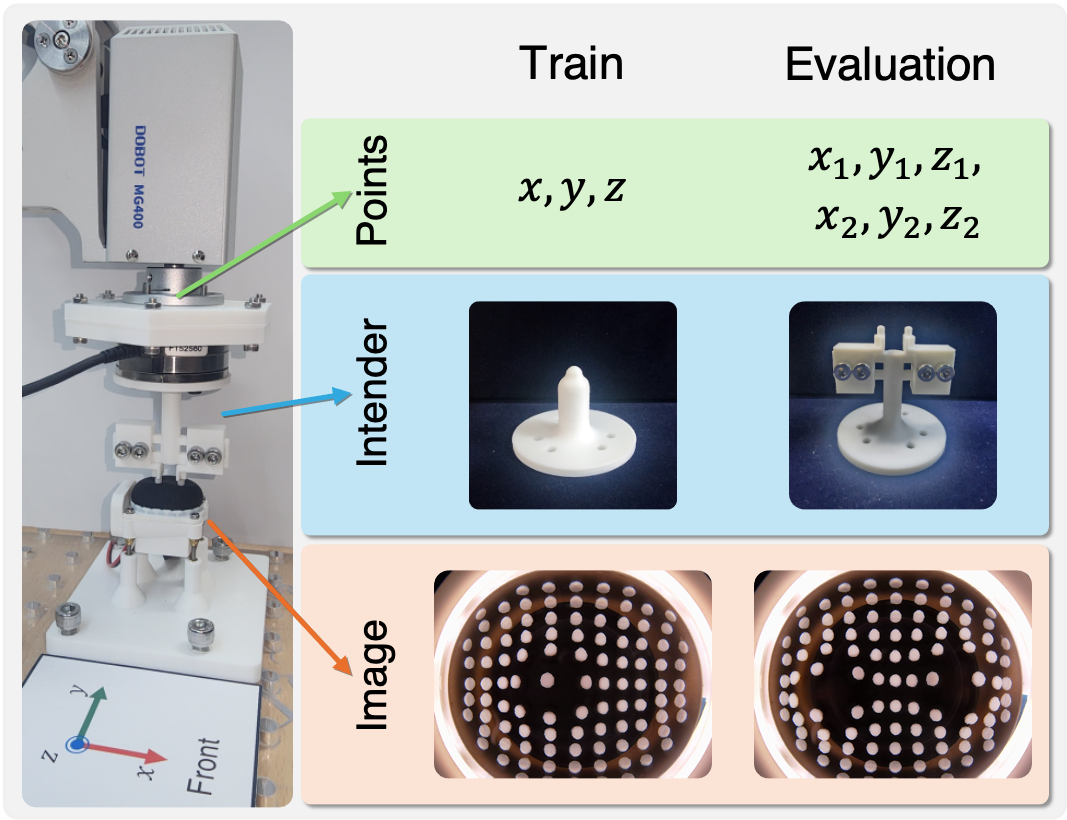}
    \caption{(Left) The experimental setup for data collection for model training. (Right) A comparison of the indenters and the corresponding data utilization for the training and evaluation of our U-Net.}
    \label{Enviroment}
    \end{center}
\end{figure}

Training TactiVerse, the U-Net-based model, necessitates a substantial dataset of input-output pairs, which we acquired using a single-point indenter (shown in the center in Fig.~\ref{Enviroment}). The left panel of Fig.~\ref{Enviroment} illustrates the experimental setup for this data acquisition. The TacTip sensor is affixed to a base, while a robotic arm, equipped with the single-point indenter, repeatedly pokes TacTip's sensing surface at randomized locations~\cite{Ward18, Nam26_TacFinRay}. Poking depth is valid only when an indentation is beyond the sensor's sensing surface. Thus, the indentation depth (-$z$ direction in Fig.~\ref{Enviroment}) is measured relative to the corresponding surface to each allocated location. We generated a set of 5,000 random 3D target points within the sensor's operational workspace, spanning a depth range of 0.5 to 6.0~mm from the sensing surface. The robotic arm then sequentially executed these indentations~\cite{Nam24RoboSoft_TacTip}.


\begin{figure}[b!]
    \begin{center}
    \includegraphics[width = 1\columnwidth]{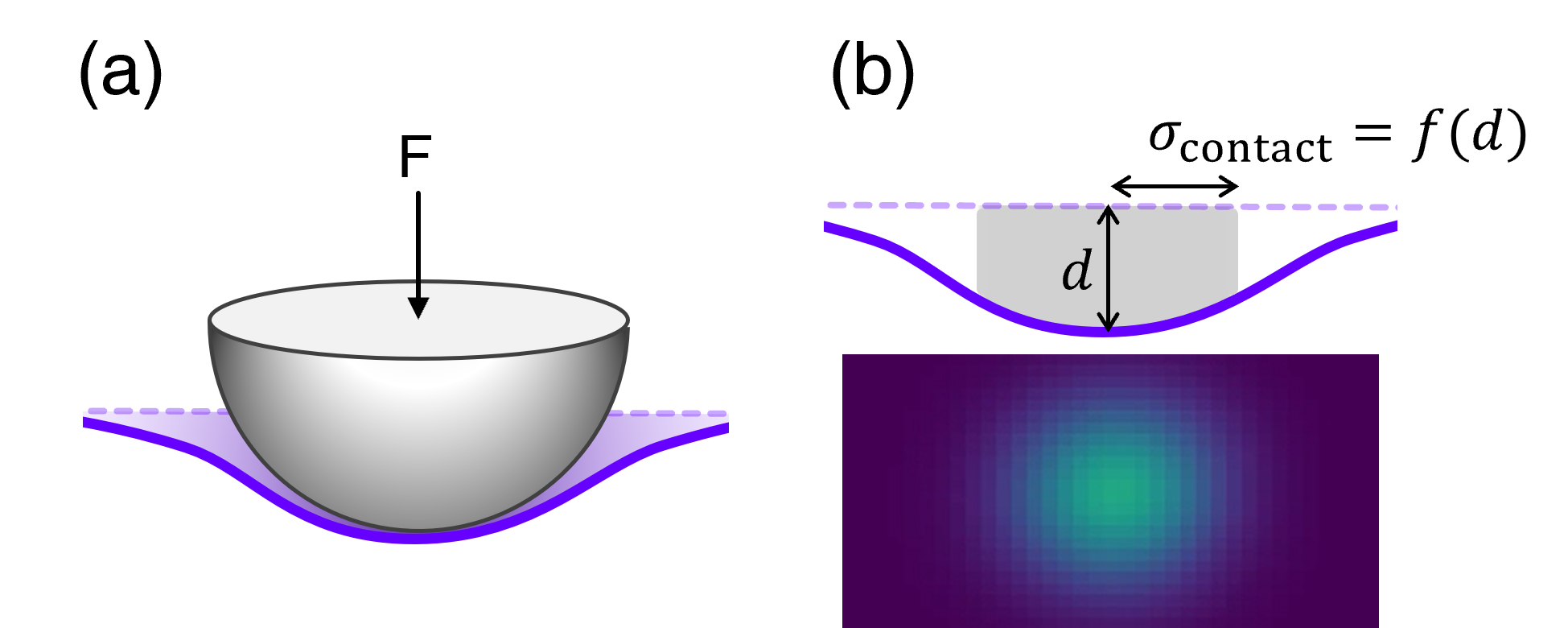}
    \caption{Design of the Gaussian kernel for depth-weighted heatmap transformation. (a) The deformation generated by a single-point indenter pressing the sensor surface is assumed to follow a 3D Gaussian kernel distribution. (b) By defining $d$ and $\sigma$ representing the kernel's size in the depth-weighted heatmap (top), a well-trained U-Net can accurately represent the actual deformation (bottom).}
    \label{Hertzian}
    \end{center}
\end{figure}

\begin{figure*}[t!]
    \centering
    \includegraphics[width = 0.8\textwidth]{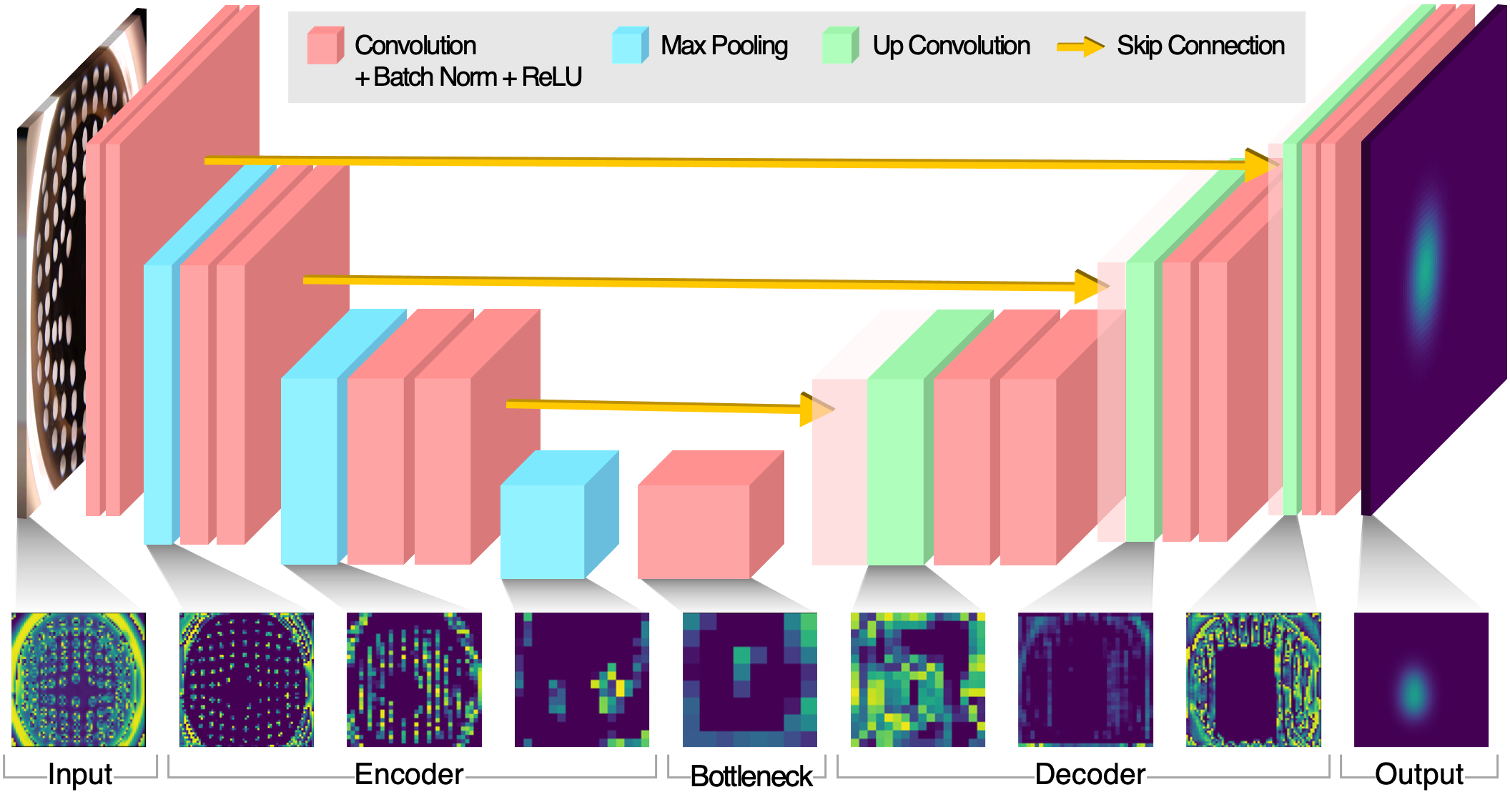}
    \caption{Overview of the U-Net architecture and an example of a feature map. The feature map visualizes the flow of information through the neural network, from input image processing in the encoder to heatmap reconstruction in the decoder.}
    \label{unet}
\end{figure*}

\subsection{Methods}
\subsubsection{Data Preprocessing}

The final output of the trained U-Net is intended to be a 3D heatmap representing the deformed geometry of the sensor surface. Therefore, we processed the 3D indentation information from the single-point training data, in conjunction with a Gaussian kernel, to generate ground-truth heatmaps analogous to the one shown in the bottom panel of Fig.~\ref{Hertzian}(b).

First, we defined a Gaussian kernel whose properties scale proportionally with the indenter's poking depth. As depicted in the top diagram of Fig.~\ref{Hertzian}(b), defining this kernel requires two parameters: its height ($d$) and its width ($\sigma_\text{contact}$). The kernel's width was designed to reflect both the indenter's contact area and the material-specific compliance of the sensor surface. We employed Hertzian contact theory to approximate the contact area, where the physical contact radius $a$ resulting from a hemispherical indenter with radius $R$ pressing an elastic surface is a function of the indentation depth $d$ ($a = \sqrt{R \cdot d}$). For our indenter, $R = 3.0$~mm. Assuming the edge of the circular deformation on the perpendicularly indented surface corresponds to $\pm3\sigma$, the kernel's standard deviation is expressed as $\sigma_{\mathrm{contact}} = a / 3$. However, as Hertzian theory assumes an ideal elastic pressure distribution, it does not fully capture the unique material compliance of the TacTip surface. Therefore, we introduced a corrected width, $\sigma_{\mathrm{final}} = \sqrt{\sigma^2_{\mathrm{contact}} + \sigma^2_{\mathrm{blur}}}$ (where we set $\sigma_{\mathrm{blur}}$ to $2.0$~mm) to ensure the kernel sufficiently reflects the actual deformation scale produced by the indenter.

The output heatmap, as shown in the bottom panel of Fig.~\ref{Hertzian}(b), has a value range normalized between 0 and 1. Consequently, we normalized the depth parameter $d$ (as $d/d_{\mathrm{max}}$). Given that the maximum indentation depth used during data collection was 6.0~mm, a heatmap value of 1 is interpreted as corresponding to this 6.0~mm physical depth.

In summary, the training data for our U-Net model (TactiVerse) consists of input RGB images, which capture the deformation of the TacTip's internal marker array, paired with output ground-truth heatmaps, each representing a single Gaussian kernel as defined above.

\subsubsection{Design of U-Net}

As illustrated by the architecture in Fig.~\ref{unet}, the U-Net model processes an input RGB image containing the marker array's deformation. The encoder path progressively compresses this image to capture high-level, abstract features related to the indentation within its bottleneck. Subsequently, the decoder path, augmented by skip connections, reconstructs a spatial heatmap that estimates the indentation's geometry and location. This design contrasts with conventional CNN models typically used for TacTip~\cite{Susini25RAL_Softness}, which often regress the output to a low-dimensional vector, such as a $3\times1$ vector representing the 3D indentation position. That approach faces significant challenges when scaling to multi-point sensing, as it requires a redesign of the output layer to accommodate the variable dimensions of the output data.

The U-Net architecture, however, inherently bypasses this output dimensionality problem by producing a pixel-wise heatmap. It enables high-quality estimation of multi-point contacts by effectively utilizing both the contextual bottleneck features and the high-resolution encoder features relayed through skip connections. The final decoder block (d1) employs a $1\times1$ convolution layer followed by a sigmoid activation function, yielding a single-channel $64\times64$ heatmap with values normalized between 0 and 1 (see Fig.~\ref{unet}).

\subsubsection{Height Scaling}

The resulting heatmap from the U-Net may contain multiple local maxima, necessitating an analysis of each maximum's location and corresponding value. To identify these contact points, we first apply a multidimensional maximum filter across the entire heatmap to pinpoint the pixel coordinates of these local peaks. Furthermore, since the pixel intensity at each maximum (normalized between 0 and 1) is proportional to the physical indentation depth (ranging from 0.5 to 6.0~mm), we can estimate the depth of each indentation by applying a scaling factor.

\subsubsection{Hyperparameters}


The model was implemented and trained using the PyTorch framework on an NVIDIA RTX 5070 Ti GPU. The hyperparameters are summarized in Table~\ref{tab:implementation_details}. The entire dataset was split into training and test sets at an 8:2 ratio, comprising 4,000 and 1,000 images, respectively. We trained the model for 50 epochs with a batch size of 8, employing the Adam optimizer with a learning rate of $3 \times 10^{-4}$. To enable the model to accurately reconstruct the ground-truth heatmaps, we applied the Binary Cross-Entropy (BCE) loss. This is well-suited for our heatmap estimation task as it effectively treats the heatmap as a pixel-wise probability distribution and handles the inherent imbalance between sparse contact regions and the background.




\begin{table}[h!]
\centering
\caption{Implementation details and hyperparameters}
\label{tab:implementation_details}
\begin{tabular}{ll}
\toprule
\textbf{Parameter} & \textbf{Value} \\
\midrule
Framework & PyTorch \\
Hardware & NVIDIA RTX 5070 Ti GPU \\
Optimizer & Adam \\
Learning Rate & $3 \times 10^{-4}$ \\
Batch Size & 8 \\
Total Epochs & 50 (with early stopping) \\
Loss Function & Binary Cross Entropy \\
Data Split & Train:Test = 4000:1000 \\
\bottomrule
\end{tabular}
\end{table}


\section{Results and Discussion}


\subsection{Exp. 1: Single-Point Sensing Comparison with Baseline}


Although our proposed U-Net architecture is primarily focused on multi-point sensing, we also conducted a comparative performance analysis against a baseline CNN model, which is commonly used for single-point sensing. The results are summarized in Table~\ref{tab:performance_comparison}. This comparison was performed on a test set of 1,000 pairs of data from the total 5,000-point single-indentation dataset.

\begin{table}[h!]
\centering
\caption{Comparative analysis of two models for predicting single-point indentation (CNN vs. U-Net)}
\label{tab:performance_comparison}
\begin{tabular}{llcccc}
\toprule
\textbf{Model} & \textbf{Target} & \textbf{$R^2$} & \textbf{MAE$^*$} & \textbf{RMSE$^*$} \\
\midrule
\multirow{4}{*}{CNN} 
& Average & 0.9965 & 0.0612 & 0.0837 \\
& X-axis   & 0.9994 & 0.0713 & 0.0908 \\
& Y-axis   & 0.9993 & 0.0732 & 0.0982 \\
& Z-axis   & 0.9908 & 0.0392 & 0.0558 \\
\midrule
\multirow{4}{*}{U-Net} 
& Average & 0.9972 & 0.0589 & 0.1282 \\
& X-axis   & 0.9982 & 0.0591 & 0.1637 \\
& Y-axis   & 0.9985 & 0.0491 & 0.1270 \\
& Z-axis   & 0.9949 & 0.0685 & 0.0938 \\
\bottomrule
\multicolumn{5}{l}{\footnotesize *Unit:\,mm}
\end{tabular}
\end{table}

In terms of averaged MAE, the U-Net (0.0589\,mm) demonstrated superior accuracy compared to the baseline CNN (0.0612\,mm). This result is noteworthy, as the U-Net's heatmap output ($64\times64$ pixels) has a discrete spatial resolution, yet it still outperforms the regression-based CNN in predicting the indentation location. However, when comparing the Root Mean Squared Error (RMSE), the U-Net's averaged RMSE (0.1282~mm) was notably higher than that of the CNN (0.0837~mm). We attribute this discrepancy to the RMSE metric's higher sensitivity to occasional large errors (outliers) compared to MAE, suggesting the U-Net may produce rare but significant miscalculations not captured by the averaged MAE. Collectively, these results indicate that the U-Net's performance is largely comparable to that of the conventional CNN for the single-point sensing task.

\subsection{Exp. 2: Dual-Point Indentation Test}
To evaluate the performance of the U-Net model on multi-point indentation tasks, despite being trained solely on single-point data, we conducted a dual-point indentation experiment. As shown in the rightmost column of Fig.~\ref{Enviroment}, we designed a distance-variable indenter and progressively reduced the distance between the two indenters from 12.0~mm down to 6.5~mm in 0.5~mm increments. This process yielded 12 distinct variables.

\begin{figure}[tbp]
    \begin{center}
    \includegraphics[width = 1\columnwidth]{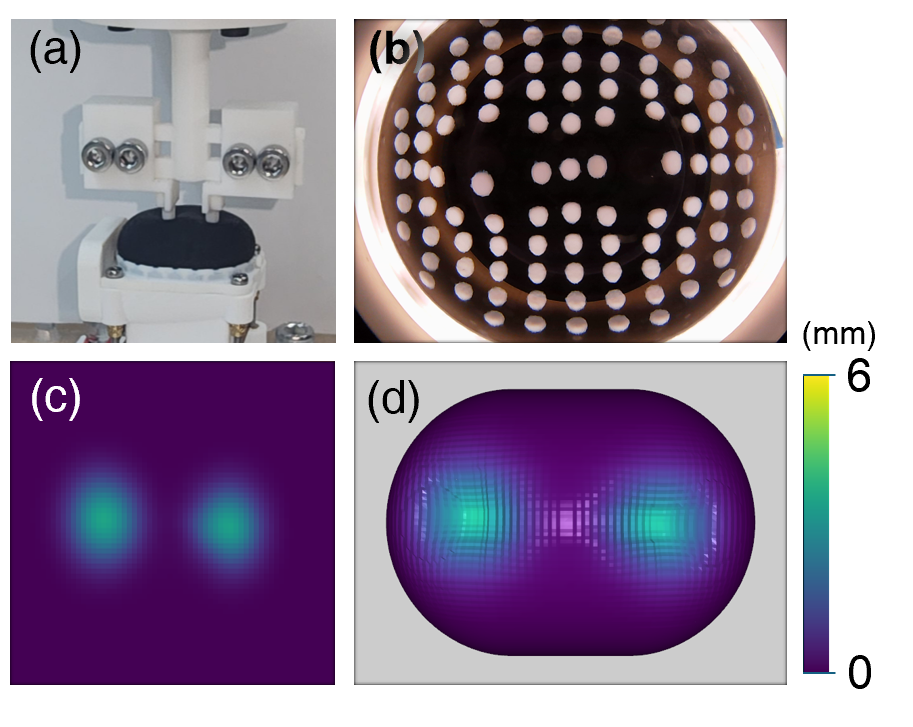}
    \caption{(a) Indentation by the robotic arm equipped with the dual-point indenter. (b) Internal marker array movement corresponding to the dual-point indentation. (c) Heatmap output generated by the trained U-Net. (d) 3D mapping of the resulting heatmap.}
    \label{Exp}
    \end{center}
\end{figure}

For each of these 12 distance conditions, the robotic arm repeatedly executed indentations at various positions across the TacTip's sensing area (Fig.~\ref{Exp}(a,\,b)). We defined the ground-truth values as the 3D positions of the two indenter tips and the predictions as the vertex locations of the Gaussian kernels generated by the U-Net's output heatmap (Fig.~\ref{Exp}(c)). Due to the sensor's inherent surface curvature, some trials resulted in only a single point of contact. These data points were filtered and excluded from our evaluation. This process resulted in a final, valid dataset of 725 dual-point contact pairs for analysis.

The blue data points in Fig.~\ref{distance} represent the results of the two-point discrimination test. Overall, the MAE decreases as the inter-point distance increases. However, an average estimation error of approximately 1.214\,mm persists regardless of the distance.

\subsubsection{Two-Point discrimination: Multi-Point model vs. Single-Point baseline}

\begin{figure}[tbp]
    \begin{center}
    \includegraphics[width = 1\columnwidth]{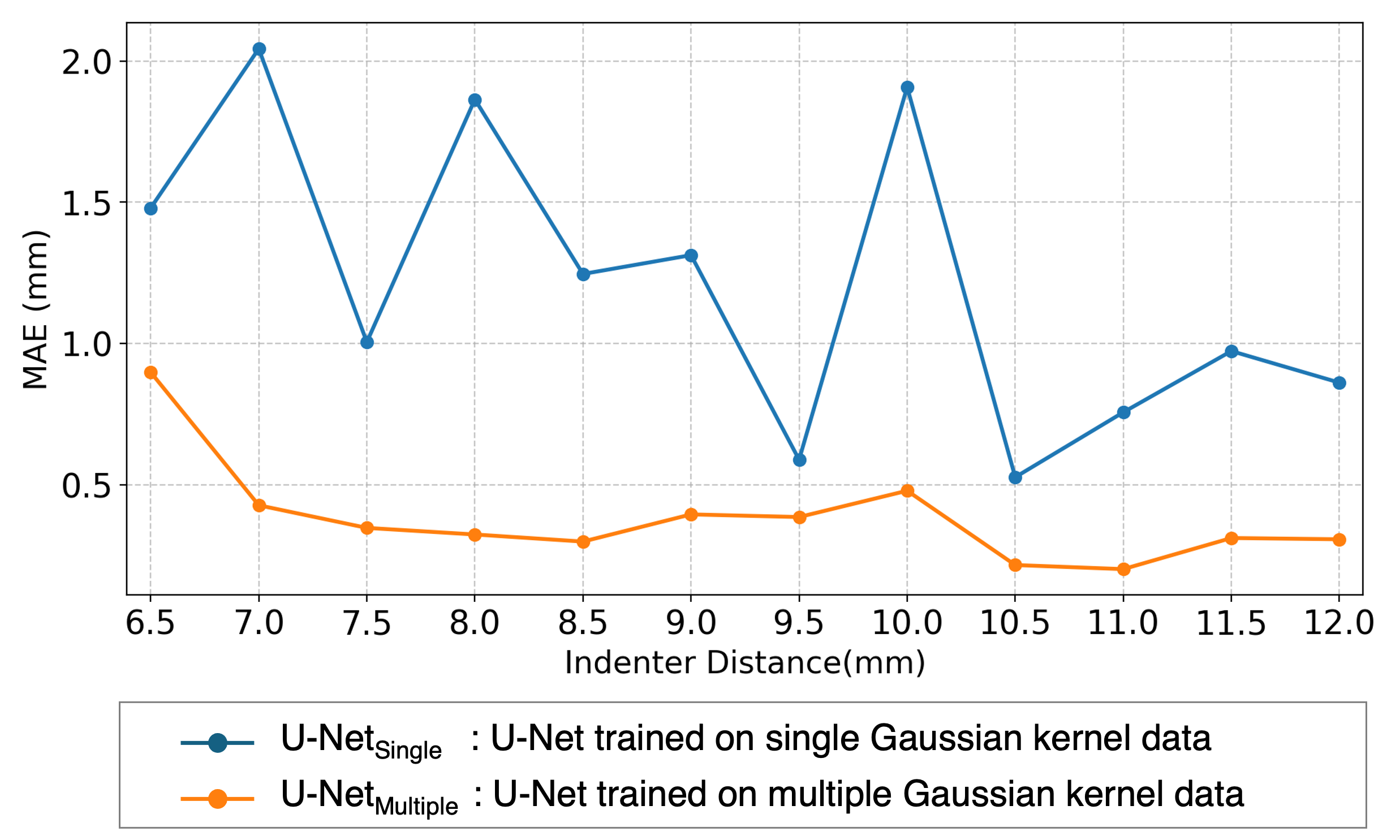}
    \caption{Mean absolute error of distance estimation by two U-Net models trained on single and multiple Gaussian kernels, respectively.}
    \label{distance}
    \end{center}
\end{figure}

We additionally evaluated an identically structured U-Net model trained on an augmented dataset featuring multiple Gaussian kernels (\textit{i.e.}, U-Net$_\text{Multiple}$). Specifically, the training set was expanded by adding 1,000 pairs each of dual- and triple-point contact data to the existing single-point dataset. To construct this dataset, we mounted dual-point (Fig.~\ref{Exp}(a)) and triple-point (Fig.~\ref{fig:threepoint}(c)) indenters on a robotic arm to collect raw images capturing the marker array displacements. The training data was then generated by applying Gaussian kernels to the corresponding indentation locations in each image. The two-point discrimination results for this new model, denoted by the orange data points in Fig.~\ref{distance}, demonstrate consistently lower MAE values across all inter-point distances compared to the baseline model (\textit{i.e.}, U-Net$_\text{Single}$) trained exclusively on single Gaussian kernels. Furthermore, the overall mean MAE significantly improved to 0.383\,mm, down from the 1.214\,mm of the baseline model. In conclusion, augmenting the training dataset with multi-point contact data substantially enhances multi-point sensing performance compared to relying solely on single-point data.

\subsection{Exp. 3: Accuracy of Position and Depth Estimation}
\subsubsection{Dual-point indenter}

In parallel with comparing the inter-point distances, we also evaluated the U-Net$_\text{Single}$'s predictions for the individual ground-truth position and depth of each indenter against the corresponding Gaussian peak information. The results are summarized in Table~\ref{tab:unet_multi_point_results}. The positional errors produced by the U-Net$_\text{Single}$ model are minimal, especially considering the R=3.0~mm radius of the indenter used for training, which demonstrates the model's excellent performance.

\begin{table}[ht]
\centering
\caption{Position and depth errors of U-Net$_\text{Single}$ for each indenter during the two-point discrimination test}
\label{tab:unet_multi_point_results}
\begin{tabular}{lc}
\toprule
Metric & Value (mm) \\
\midrule
Mean positional error (mm) & 2.24 \\
MAE in depth (mm) & 0.71 \\
\bottomrule
\end{tabular}
\end{table}

\subsubsection{Triple-point indenter}

\begin{figure}[tbp]
    \centering
    \includegraphics[width=0.9\columnwidth]{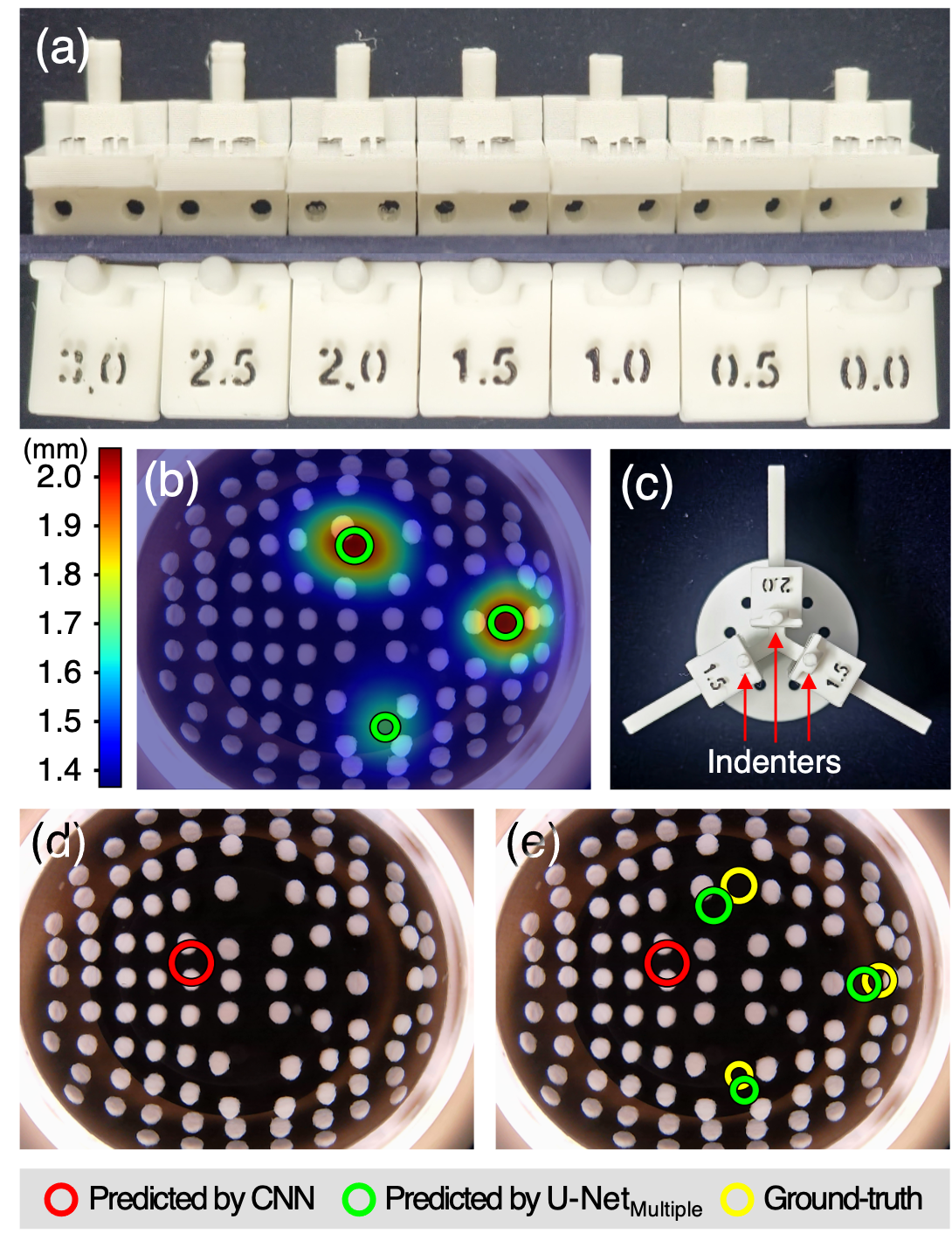}
    \caption{Visualization of predictions by the proposed U-Net and the single-point regression CNN baseline in triple-point contact scenarios with varying depths. (a) Modular indenters with varying heights in 0.5\,mm increments ($h$: 0.0 to 3.0\,mm), which can be inserted into the triple-point indenter set in various combinations. (b) Overlay of the deformation heatmap generated by U-Net$\text{Multiple}$ and the estimated positions extracted from the heatmap peaks during a triple-point indentation on the TacTip. (c) The assembled triple-point indenter set. (d) Example output of the CNN baseline, which is restricted to single-point prediction and thus infers only one location despite the three-point contact. (e) Comparison of the estimated positions by the CNN baseline and U-Net$\text{Multiple}$ against the ground-truth locations under triple-point indentation.}
    \label{fig:threepoint}
\end{figure}

Since the dual-point symmetric contact experiments involved indenters with identical depths, they did not account for relative depth variations. To evaluate the model's accuracy in estimating these relative depth differences, we conducted an additional triple-point contact experiment featuring varying indentation depths. Figure~\ref{fig:threepoint} illustrates the modular triple-point indenter configuration, which utilizes depth increments of 0.5\,mm, alongside a visual comparison of the predictions from the single-point regression CNN baseline and the proposed U-Net$_\text{Multiple}$. Furthermore, the quantitative estimation errors from this comparison, including the prediction results obtained using the U-Net$_\text{Multiple}$ model, are summarized in Table~\ref{tab:multi_validation}.



\begin{table}[tbp]
\centering
\caption{Quantitative evaluation on varying-depth multi-contact datasets (U-Net$_\text{Single}$ vs. U-Net$_\text{Multiple}$)}
\label{tab:multi_validation}
\resizebox{\columnwidth}{!}{%
\begin{tabular}{l c l c c}
\toprule
\textbf{\begin{tabular}[c]{@{}l@{}}\# of indentations \\ per input image\end{tabular}} & \textbf{\begin{tabular}[c]{@{}c@{}}Number of \\ test samples ($n$)\end{tabular}} & \textbf{\begin{tabular}[c]{@{}c@{}}Model \\ for inference\end{tabular}} & \textbf{\begin{tabular}[c]{@{}c@{}}Mean position \\ error (mm)\end{tabular}} & \textbf{\begin{tabular}[c]{@{}c@{}}Mean depth \\ error (mm)\end{tabular}} \\
\midrule
\multirow{2}{*}{Dual-point} 
& \multirow{2}{*}{908} 
& U-Net$_\text{Single}$ & 0.408 & 0.164 \\
& & U-Net$_\text{Multiple}$ & 0.412 & 0.167 \\
\midrule
\multirow{2}{*}{Triple-point} 
& \multirow{2}{*}{1025} 
& U-Net$_\text{Single}$ & 0.526 & 0.178 \\
& & U-Net$_\text{Multiple}$ & 0.500 & 0.178 \\
\bottomrule
\end{tabular}
} %
\end{table}

\noindent 1) \textbf{Mean position error}:
The mean position error was calculated using the 2D Euclidean distance, excluding the depth estimation error. The results indicate that both models (U-Net$_\text{Single}$ and U-Net$_\text{Multiple}$) yielded an error of approximately 0.4\,mm when tested on dual-point images, and approximately 0.5\,mm on triple-point images. In both cases, no significant difference was observed between the two models. This suggests that the model trained exclusively on single-point data (U-Net$_\text{Single}$) suffers no performance degradation compared to its multi-point counterpart. However, we observed that increasing the number of contact points in the input image from two to three resulted in a slight increase in the mean position error of approximately 0.1\,mm (from 0.4\,mm to 0.5\,mm).





\noindent 1) \textbf{Mean depth error}:
For the same test samples, the mean depth errors at each contact point for U-Net$_\text{Single}$ and U-Net$_\text{Multiple}$ were 0.164\,mm and 0.167\,mm in the dual-point scenario, and 0.178\,mm and 0.178\,mm in the triple-point scenario, respectively. Consistent with the mean position error results, the difference in training datasets did not lead to any significant difference in the estimation errors between the two models. Interestingly, depth estimation also exhibited a slight increase in mean error as the number of contact points increased, rising from 0.1655\,mm to 0.178\,mm.


\section{CONCLUSION}
This paper introduced TactiVerse, a U-Net framework designed to address the challenge of generalizing complex, multi-point contact scenarios in soft tactile sensors using limited training data. By effectively utilizing a dataset comprising exclusively single-point indentations, our approach enables data-driven contact geometry estimation for complex patterns.

Our evaluations demonstrated that the proposed architecture achieves highly accurate single-point sensing, yielding a superior Mean Absolute Error (MAE) of 0.0589\,mm compared to the 0.0612\,mm of the conventional CNN baseline. When applied to multi-point tasks, the model trained exclusively on single-point data suffered no performance degradation compared to its multi-point counterpart. However, augmenting the training dataset with actual multi-point contact data substantially enhanced the overall sensing capabilities, significantly improving the overall mean MAE from 1.214\,mm to 0.383\,mm. Furthermore, we observed a consistent trend where increasing the number of contact points from two to three resulted in a slight increase in both the mean position error (from 0.4\,mm to 0.5\,mm) and the mean depth error (rising from 0.1655\,mm to 0.178\,mm).

Ultimately, this methodology successfully expands the functional scope of vision-based soft sensors, enabling multi-point and complex contact pattern sensing without necessitating auxiliary hardware or imposing rigid constraints near the contact interface. This ensures that the intrinsic morphological advantages and overall compliance of the soft materials are fully preserved.

\addtolength{\textheight}{-12cm}   




\section*{ACKNOWLEDGMENT}
This work was supported by the National Research Foundation of Korea (NRF) grant funded by the Korea government (MSIT) (RS-2023-00242528, RS-2024-00436182) and by the IITP (Institute for Information \& Communications Technology Planning \& Evaluation)-ITRC (Information Technology Research Center) grant funded by the Korea government (No. IITP-2026-RS-2024-00437756).

\bibliography{ref}

\end{document}